\documentclass{article}
\usepackage{amsmath,graphicx,spconf}
\usepackage{mathtools} 
\usepackage{amssymb}
\usepackage{tikz,pgfplots}
\usepackage[inline]{enumitem}
\usepackage{xspace} 
\usepackage{bm,bbm}
\usetikzlibrary{matrix,positioning,arrows}
\usetikzlibrary{decorations.markings}
\usetikzlibrary{calc}
\usetikzlibrary{external}
\pgfplotsset{compat=1.13}
\usepackage{nameref}
\usepackage{hyperref}
\hypersetup{%
  colorlinks   = true,
  citecolor    = blue  
}

\let\oldsec\section
\renewcommand{\section}[1]{\vspace{-5pt}\oldsec{#1}\vspace{-5pt}}
\let\oldsubsec\subsection
\renewcommand{\subsection}[1]{\vspace{-5pt}\oldsubsec{#1}\vspace{-5pt}}

\toappear{To Appear in the 27th IEEE International Workshop on Machine Learning
For Signal Processing (MLSP) 2017}

\newcommand{\CWT}{\ensuremath{{\mathbb{C}}\mathrm{WT}}\xspace} 
\newcommand{\DTCWT}{{\ensuremath{\mathrm{DT}\CWT}}\xspace}

\newcommand{\x}{\times}                     
\newcommand{\degs}{{\ensuremath{^{\circ}}}\xspace}
\newcommand{\conv}{\ast}
\newcommand{\definedas}{\triangleq}

\newcommand{\reals}[1][]{\ensuremath{{\mathbb{R}}^{#1}}\xspace}
\newcommand{\complexes}[1][]{\ensuremath{{\mathbb{C}}^{#1}}\xspace}

\usepackage{silence}

\title{Visualizing and Improving Scattering Networks}
\name{Fergal Cotter and Nick Kingsbury}
\address{Signal Processing Group, Department of Engineering, University of
Cambridge, U.K.}
\begin{document}
%

\maketitle
\begin{abstract}
  Scattering Transforms (or ScatterNets) introduced by Mallat in
  \cite{mallat_group_2012} are a promising start into creating a well-defined
  feature extractor to use for pattern recognition and image classification
  tasks. They are of particular interest due to their architectural similarity
  to Convolutional Neural Networks (CNNs), while requiring no parameter learning
  and still performing very well (particularly in constrained classification
  tasks). 
  
  In this paper we visualize what the deeper layers of a ScatterNet 
  are sensitive to using a `DeScatterNet'. We show that the higher orders of ScatterNets are sensitive
  to complex, edge-like patterns (checker-boards and rippled edges). These
  complex patterns may be useful for texture classification, but are quite
  dissimilar from the patterns visualized in second and third layers of
  Convolutional Neural Networks (CNNs) - the current state of the art Image
  Classifiers. We propose that this may be the source of the current gaps in
  performance between ScatterNets and CNNs (83\% vs 93\% on CIFAR-10 for
  ScatterNet+SVM vs ResNet).
  We then use these visualization tools to propose possible enhancements to the
  ScatterNet design, which show they have the power to extract features more
  closely resembling CNNs, while still being well-defined and having the
  invariance properties fundamental to ScatterNets. 

\end{abstract}
\section{Introduction}
\label{sec:intro}
Scattering transforms, or ScatterNets, have recently gained much attention and
use due to their ability to extract generic and descriptive features in well
defined way. They can be used as unsupervised feature extractors
for image classification \cite{bruna_invariant_2013, oyallon_deep_2015, 
singh_dual-tree_2017, singh_multi-resolution_2017} and texture classification
\cite{sifre_rotation_2013}, or in combination with supervised methods such as
Convolutional Neural Networks (CNNs) to make the latter learn quicker, and in
a more stable way \cite{oyallon_scaling_2017}. 

ScatterNets have been shown to perform very well as image classifiers. In
particular, they can outperform CNNs for classification tasks with reduced
training set sizes, e.g.\ in CIFAR-10 and CIFAR-100 (Table 6 from
\cite{oyallon_scaling_2017} and Table 4 from \cite{singh_dual-tree_2017}).  
They are also near state-of-the-art for Texture Discrimination tasks
(Tables 1--3 from \cite{sifre_rotation_2013}). Despite this, there still exists
a considerable gap between them and CNNs on challenges like CIFAR-10 with the
full training set ($~83\%$ vs.\ $~93\%$). Even considering the benefits of
ScatterNets, this gap must be addressed.

We first revise the operations that form a ScatterNet in
\autoref{sec:scatternet}. We then introduce our DeScatterNet
(\autoref{sec:descatternet}), and show how we can use it to examine the layers of ScatterNets
(using a similar technique to the CNN visualization in
\cite{zeiler_visualizing_2014}). We use this analysis tool to
highlight what patterns a ScatterNet is sensitive to
(\autoref{sec:visualization}), showing that they are very
different from what their CNN counterparts are sensitive to, and possibly less
useful for discriminative tasks. 

We use these observations to propose an architectural change to ScatterNets,
which have not changed much since their inception in \cite{mallat_group_2012}. 
Two changes of note however are the work of Sifre and Mallat in
\cite{sifre_rotation_2013}, and the work of Singh and Kingsbury in
\cite{singh_dual-tree_2017}.  Sifre and Mallat introduced Rotationally Invariant
ScatterNets which took ScatterNets in a new direction, as the architecture now
included filtering across the wavelet orientations (albeit with heavy
restrictions on the fitlers used).  Singh and Kingsbury achieved improvements in
performance in a Scattering system using the spatially implementable \DTCWT
\cite{kingsbury_complex_2001} wavelets instead of the Fourier Transform (FFT)
based Morlet previously used.

We build on these two systems, showing that with carefully designed complex
filters applied across the complex spatial coefficients of a 2-D \DTCWT,  
we can build filters that are sensitive to more recognizable shapes like
those commonly seen in CNNs, such as corners and curves (\autoref{sec:corners}). 

\section{The Scattering Transform}
\label{sec:scatternet}

The Scattering Transform, or ScatterNet, is a cascade of complex wavelet transforms and
modulus non-linearities (throwing away the phase of the complex wavelet
coefficients). At a chosen scale, averaging filters provide invariance
to nuisance variations such as shift and deformation (and potentially
rotations). Due to the non-expansive nature of the wavelet
transform and the modulus operation, this transform is stable to
deformations. 

Typical implementations of the ScatterNet are limited to two `orders'
(equivalent to layers in a CNN) \cite{oyallon_deep_2015, singh_dual-tree_2017,
oyallon_scaling_2017}. In addition to scattering \emph{order}, we also have the
\emph{scale} of invariance, $J$. This is the number of band-pass coefficients
output from a wavelet filter bank (FB), and defines the cut-off frequency for
the final low-pass output: $2^{-J}\frac{f_s}{2}$ ($f_s$ is the sampling
frequency of the signal). Finally, we call the number of oriented wavelet
coefficients used $L$.  These are the three main hyper-parameters of the
scattering transform and must be set ahead of time. We describe a system with
scale parameter $J=4$, order $m=2$ and with $L=6$ orientations ($L$ is fixed
to 6 for the \DTCWT but is flexible for the FFT based Morlet wavelets).

Consider an input signal $x(\bm{u}), \bm{u}\in\reals[2]$. The zeroth order
\textbf{scatter} coefficient is the lowpass output of a $J$ level FB: 

\begin{equation}
  S_0x(\bm{u}) \definedas (x \conv \phi_J)(\bm{u})
\end{equation}

This is invariant to translations of up to $2^J$ pixels\footnote{From here on,
we drop the $\bm{u}$ notation when indexing $x$, for clarity.}. In exchange for
gaining invariance, the $S_0$ coefficients have lost a lot of information
(contained in the rest of the frequency space). The remaining energy of $x$ is
contained within the first order \textbf{wavelet} coefficients:

\begin{equation}
W_1x(\bm{u}, j_1, \theta_1) \definedas x \conv \psi_{j_1, \theta_1}
\end{equation}

for $j_1\in\{1,2, \ldots, J\}, \theta_1\in\{1,2, \ldots, L\}$. We will want to
retain this information in these coefficients to build a useful classifier.

Let us call the set
of available scales and orientations $\Lambda_1$ and use $\lambda_1$ to index it.
For both Morlet and \DTCWT\ implementations, $\psi$ is complex-valued, i.e.,
$\psi = \psi^r + j\psi^i$ with $\psi_r$ and $\psi_i$ forming a Hilbert Pair,
resulting in an analytic $\psi$.
 This analyticity provides a source of
invariance --- small input shifts in $x$ result in a phase rotation (but little
magnitude change) of the complex wavelet coefficients\footnote{In comparison to
a system with purely real filters such as a CNN, which would have rapidly
varying coefficients for small input shifts \cite{kingsbury_complex_2001}.}.

Taking the magnitude of $W_1$ gives us the first order \textbf{propagated}
signals:

\begin{equation}
  U_1x(\bm{u}, \lambda_1) \definedas |x\conv \psi_{\lambda}| 
    = \sqrt{(x \conv \psi^r_{\lambda_1})^2 + (x \conv \psi^i_{\lambda_1})^2}
\end{equation}

The first order scattering coefficient makes $U_1$ invariant up to our
scale $J$ by averaging it:

\begin{equation}
  S_1x(\bm{u}, \lambda_1) \definedas |x\conv \psi_{\lambda}| \conv \phi_J
\end{equation}

If we define $U_0 \definedas x$, then we can iteratively define:

\begin{eqnarray}
  W_{m} &=& U_{m-1} \conv \psi_{\lambda_{m}} \\
  U_{m} &=& |W_{m}| \\
  S_{m} &=& U_m \conv \phi_J
\end{eqnarray}

We repeat this for higher orders, although previous work shows that, for natural
images, we get diminishing returns after $m=2$. The output of our ScatterNet is
then:

\begin{equation}
  Sx = \{S_0x, S_1x, S_2x\}
\end{equation}

\subsection{Scattering Color Images}
A wavelet transform like the \DTCWT\ accepts single channel input, while we
often work on RGB images. This leaves us with a choice. We can either:
\begin{enumerate}
  \item Apply the wavelet transform (and the subsequent scattering operations)
    on each channel independently. This would triple the output size to $3C$.
  \item Define a frequency threshold below which we keep color information, and
    above which, we combine the three channels into a single luminance channel.
\end{enumerate}
The second option uses the well known fact that the human eye is far less sensitive 
to higher spatial frequencies in color channels than in luminance channels. 
This also fits in with the first layer filters seen in the well known
Convolutional Neural Network, AlexNet. Roughly one half of the filters were low
frequency color `blobs', while the other half were higher frequency, grayscale,
oriented wavelets. 

For this reason, we choose the second option for the
architecture described in this paper. We keep the 3 color
channels in our $S_0$ coefficients, but work only on grayscale for high orders 
(the $S_0$ coefficients are the lowpass bands of a J-scale wavelet transform, so
we have effectively chosen a color cut-off frequency of $2^{-J} \frac{f_s}{2}$).

For example, consider an RGB input image $x$ of size $64\x 64\x 3$. The 
scattering transform we have described with parameters $J=4$ and $m=2$ would
then have the following coefficients:
\begin{eqnarray*}
  S_0 &:& (64\x 2^{-J}) \x (64\x 2^{-J}) \x 3 = 4\x 4\x 3 \\
  S_1 &:& 4\x 4\x (L J) = 4\x 4\x 24 \\
  S_2 &:& 4\x 4\x \left(\frac{1}{2}L^2 J(J-1)\right) = 4\x 4\x 216 \\
  S &:& 4\x 4\x (216+24+3) = 4\x 4\x 243
\end{eqnarray*}

\section{The Inverse Network}
\label{sec:descatternet}
\begin{figure*}[ht]
  \centering
  \includegraphics[width=\textwidth]{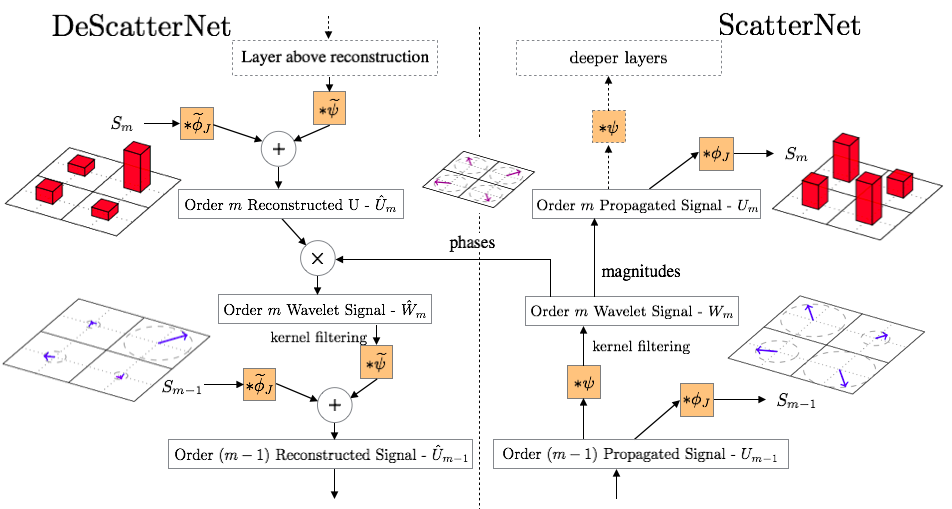}
  \caption{A DeScattering layer (left) attached to a Scattering layer (right).
  We are using the same convention as \cite{zeiler_visualizing_2014}
  Figure 1 - the input signal starts in the bottom right hand corner, passes
  forwards through the ScatterNet (up the right half), and then is 
  reconstructed in the DeScatterNet (downwards on the left half).
  The DeScattering layer will reconstruct an approximate version of the previous
  order's propagated signal. The $2\x 2$ grids shown around the image are either Argand
  diagrams representing the magnitude and phase of small regions of
  \emph{complex} (De)ScatterNet
  coefficients, or bar charts showing the magnitude of the \emph{real}
  (De)ScatterNet coefficients (after applying the modulus non-linearity). For reconstruction, we
  need to save the discarded phase information and reintroduce it by multiplying
  it with the reconstructed magnitudes.}
  \label{fig:descat}
\end{figure*}
\label{sec:invscat}
We now introduce our inverse scattering network. This allows us to back project
scattering coefficients to the image plane; it is inspired by the
\textbf{DeconvNet} used by Zeiler and Fergus in
\cite{zeiler_visualizing_2014} to look into the deeper layers of CNNs. 

We emphasize that instead of thinking about perfectly reconstructing $x$ from
$S\in \reals[H'\x W'\x C]$, we want to see what signal/pattern in the input image caused
a large activation in each channel. This gives us a good idea of what each
output channel is sensitive to, or what it extracts from the input. 
Note that we do not use any of the log normalization layers described in
\cite{oyallon_deep_2015, singh_dual-tree_2017}.

\subsection{Inverting the Low-Pass Filtering}
Going from the $U$ coefficients to the $S$ coefficients involved convolving by
a low pass filter, $\phi_J$ followed by decimation to make the output $(H\x
2^{-J})\x (W\x2^{-J})$.  $\phi_J$ is a purely real filter, and we can `invert'
this operation by interpolating $S$ to the same spatial size as $U$ and convolving with
the mirror image of $\phi_J$, $\widetilde{\phi}_J$ (this is equivalent to the
transpose convolution described in \cite{zeiler_visualizing_2014}). 

\begin{equation}
  \label{eq:s_hat}
  \hat{S}_{m} = S_{m} \conv \widetilde{\phi}_J
\end{equation}

This will not recover $U$ as it was on the forward pass, but will recover all
the information in $U$ that caused a strong response in $S$.

\subsection{Inverting the Magnitude Operation}
In the same vein as \cite{zeiler_visualizing_2014}, we face a difficult
task in inverting the non-linearity in our system. 
We lend inspiration from the \emph{switches} introduced in the DeconvNet; the
switches in a DeconvNet save the location of maximal activations so that
on the backwards pass activation layers could be unpooled trivially. We do an
equivalent operation by saving the phase of the complex activations. On the
backwards pass we reinsert the phase to give our recovered $W$.
\begin{equation}
  \label{eq:w_hat}
  \hat{W}_{m} = \hat{U}_{m}e^{j\theta_{m}}
\end{equation}

\subsection{Inverting the Wavelet Decomposition}
Using the \DTCWT makes inverting the wavelet transform simple, as we
can simply feed the coefficients through the synthesis filter banks to regenerate
the signal. For complex $\psi$, this is convolving with the conjugate transpose
$\widetilde{\psi}$: 

\begin{eqnarray}
  \label{eq:x_hat}
  \hat{U}_{m-1} &=& \hat{S}_{m-1} + \hat{W}_{m} \nonumber \\
              &=& S_{m-1} \conv \widetilde{\phi}_J + \sum_{j, \theta} W_{m}(\bm{u}, j,
  \theta) \conv \widetilde{\psi}_{j, \theta}
\end{eqnarray}

\begin{figure*}[htp]
  \centering
  \includegraphics[width=0.9\textwidth]{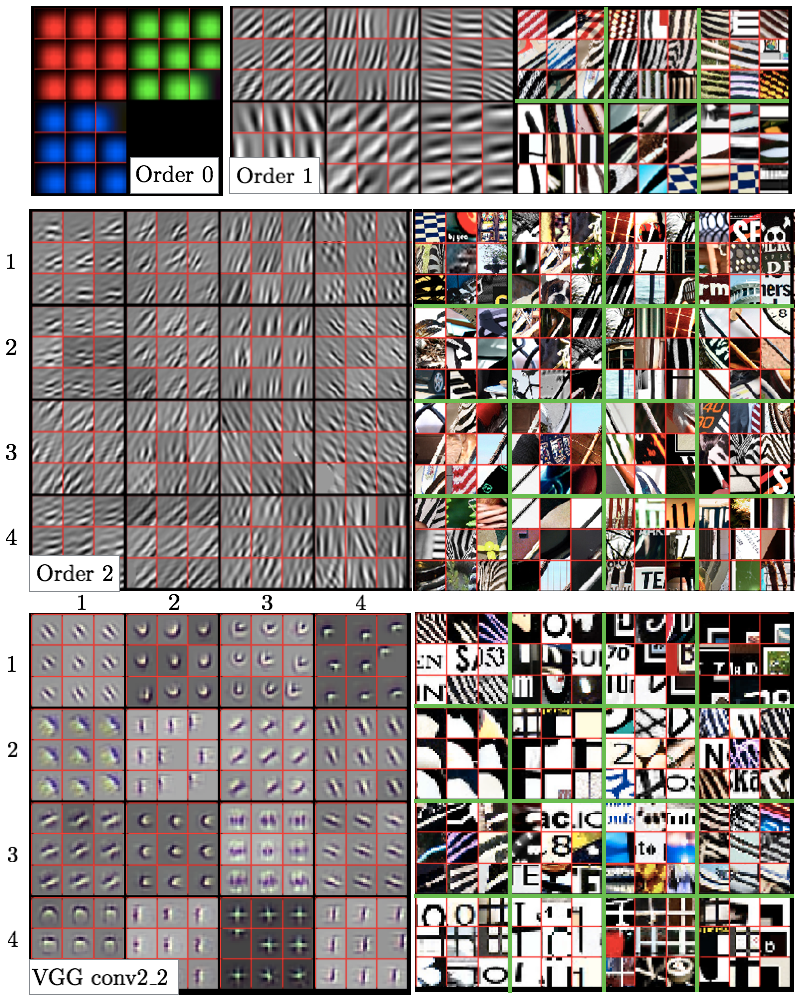}
  \caption{Visualization of a random subset of features from $S_0$ (all
  3), $S_1$ (6 from the 24) and $S_2$ (16 from the 240) scattering
    outputs. We record the top 9 activations for the chosen features and project
    them back to the pixel space. We show them alongside the input image patches
    which caused the large activations. We also include reconstructions from
    layer conv2\_2 of VGG Net \cite{simonyan_very_2014}(a popular CNN, often used
    for feature extraction) for reference --- here we display 16 of the 128
    channels. The VGG reconstructions were made with a CNN DeconvNet based on
    \cite{zeiler_visualizing_2014}. Image best viewed digitally.}
  \label{fig:reconstructions}
\end{figure*}
\begin{figure*}[t]
  \centering
  \includegraphics[width=0.9\textwidth]{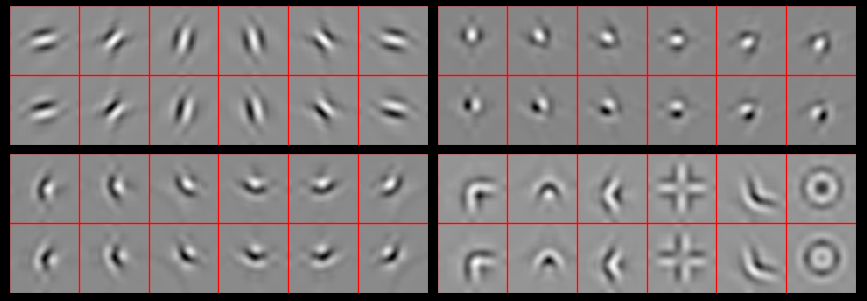}
  \caption{Shapes possible by filtering across the wavelet orientations with
  complex coefficients. All shapes are shown
  in pairs: the top image is reconstructed from a purely real output, and the
  bottom image from a purely imaginary output. These `real' and `imaginary' shapes 
  are nearly orthogonal in the pixel space (normalized dot product $<0.01$ for
  all but the doughnut shape in the bottom right, which has $0.15$) but produce
  the same $U'$, something that would not be possible without the complex
  filters of a ScatterNet.  Top left - reconstructions from $U_1$ (i.e.\ no
  cross-orientation filtering). Top right- reconstructions from $U_1'$ using
  a $1\x 1\x 12$ Morlet Wavelet, similar to what was done in the `Roto-Translation'
  ScatterNet described in \cite{sifre_rotation_2013, oyallon_deep_2015}. Bottom
  left - reconstructions from $U_1'$ made with a more general $1\x 1\x 12$
  filter, described in \autoref{eq:simple_corner}. Bottom
  right - some reconstructions possible by filtering a general $3\x 3\x 12$ filter. }
  \label{fig:newshapes}
\end{figure*}

\section{Visualization with Inverse Scattering}
\label{sec:visualization}

To examine our ScatterNet, we scatter all of the images from ImageNet's validation
set and record the top 9 images which most highly activate each of the $C$
channels in the ScatterNet. This is the \emph{identification} phase (in which no
inverse scattering is performed). 

Then, in the \emph{reconstruction}
phase, we load in the $9\x C$ images, and scatter them one by one. We take the
resulting $4\x 4\x 243$ output vector and mask all but a single value in the channel we are currently examining.

This 1-sparse tensor is then presented to the inverse scattering network from
\autoref{fig:descat} and projected back to the image space. Some results of this
are shown in \autoref{fig:reconstructions}. This figure shows reconstructed
features from the layers of a ScatterNet. For a given output channel, we show
the top 9 activations projected independently to pixel space. For the first and
second order coefficients, we also show the patch of pixels in the input image
which cause this large output. We display activations from various scales
(increasing from first row to last row), and random orientations in these
scales. 

The order 1 scattering (labelled with `Order 1' in
\autoref{fig:reconstructions}) coefficients look quite similar to the first
layer filters from the well known AlexNet CNN \cite{krizhevsky_imagenet_2012}.
This is not too surprising, as the first order scattering coefficients are
simply a wavelet transform followed by average pooling. They are responding to
images with strong edges aligned with the wavelet orientation. 

The second order coefficients (labelled with `Order
2' in \autoref{fig:reconstructions}) appear very similar to the order
1 coefficients at first glance.
They too are sensitive to edge-like features, and some of them (e.g.\ third row,
third column and fourth row, second column) are mostly just that. These are
features that have the same oriented wavelet applied at both the first and
second order.  Others, such as the 9 in the first row, first column, and first
row, fourth column are more sensitive to checker-board like patterns. Indeed,
these are activations where the orientation of the wavelet for the first and
second order scattering were far from each other (15\degs and 105\degs for the
first row, first column and 105\degs and 45\degs for the first row, fourth
column).

For comparison, we include reconstructions from the second layer of the
well-known VGG CNN\@ (labelled with `VGG conv2\_2', in
\autoref{fig:reconstructions}). These were made with a DeconvNet, following the
same method as \cite{zeiler_visualizing_2014}. Note that while some of
the features are edge-like, we also see higher order shapes like corners,
crosses and curves.

\section{Corners, Crosses and Curves}
\label{sec:corners}
These reconstructions show that the features extracted from ScatterNets vary
significantly from those learned in CNNs after the first order. In many
respects, the features extracted from a CNN like VGGNet look preferable for use
as part of a classification system.

\cite{sifre_rotation_2013} and \cite{oyallon_deep_2015} introduced the idea of
a `Roto-Translation' ScatterNet. Invariance to rotation could be made by
applying averaging (and bandpass) filters across the $L$ orientations 
from the wavelet transform \emph{before} applying the complex modulus.
Momentarily ignoring the form of the filters they apply, referring to them
as $F_k\in \complexes[L]$, we can think of this stage as stacking the $L$
outputs of a complex wavelet transform on top of each other, and convolving
these filters $F_k$ over all spatial locations of the wavelet coefficients $W_m
x$ (this is equivalent to how filters in a CNN are fully connected
in depth):

\begin{equation}
  V_m x(\bm{u}, j, k) = W_{m}x \conv F_k = \sum_{\theta} W_{m}x(\bm{u}, j, \theta)
F_k(\theta)
\end{equation}

We then take the modulus of these complex outputs to make a second propagated
signal:

\begin{equation}
  U_{m}'x \definedas |V_{m}x| = |W_{m}x \conv F_k| = |U_{m-1}x
  \conv \psi_{\lambda_{m}} \conv F_k|
\end{equation}

We present a variation on this idea, by filtering with a more general 
$F\in \complexes[H\x W\x 12]$. We use $F$ of length 12 rather than 6, as we use
the $L=6$ orientations and their complex conjugates; each wavelet is a 30\degs
rotation of the previous, so with 12 rotations, we can cover the full
$360\degs$. 

\autoref{fig:newshapes} shows some reconstructions from these $V$ coefficients.
Each of the four quadrants show reconstructions from a different class of
ScatterNet layer. 
All shapes are shown in real and imaginary Hilbert-like pairs; the top images in
each quadrant are reconstructed from a purely real $V$, while the bottom inputs
are reconstructed from a purely imaginary $V$. This shows one level of invariance of
these filters, as after taking the complex magnitude, both the top and the
bottom shape will activate the filter with the same strength. In comparison, for
the purely real filters of a CNN, the top shape would cause a large output, and
the bottom shape would cause near 0 activity (they are nearly orthogonal to each
other).

In the top left, we display the 6 wavelet filters for reference (these were
reconstructed from $U_1$, not $V_1$). In the top right of the figure we see some
of the shapes made by using the $F$'s from the Roto-Translation ScatterNet
\cite{sifre_rotation_2013, oyallon_deep_2015}.  The bottom left is where we
present some of our novel kernels. These are simple corner-like shapes made 
by filtering with ${F\in \complexes[1\x 1\x 12]}$
\begin{equation}
\label{eq:simple_corner}
F = [1, j, j, 1, 0, 0, 0, 0, 0, 0, 0, 0]
\end{equation}
The six orientations are made by rolling the coefficients in $F$ along one
sample (i.e. $[0, 1, j, j, 1, 0,\ldots]$, $[0,0,1,j,j,1,0,\ldots]$,
$[0,0,0,1,j,j,1,0, \ldots]$ \ldots). Coefficients roll back around (like
circular convolution) when they reach the end.

Finally, in the bottom right we see shapes made by 
${F \in \complexes[3\x 3\x 12]}$. Note that with the exception of the 
ring-like shape which has 12 non-zero coefficients, all of these shapes were
reconstructed with $F$'s that have 4 to 8 non-zero coefficients of a possible 
64. These shapes are now beginning to more closely resemble the more complex
shapes seen in the middle stages of CNNs. 

\section{Discussion}
This paper presents a way to investigate what the higher orders of a ScatterNet
are responding to - the DeScatterNet described in \autoref{sec:descatternet}.
Using this, we have shown that the second `layer' of a ScatterNet 
responds strongly to patterns that are very dissimilar to those that highly activate the
second layer of a CNN\@. As well as being dissimilar to CNNs, visual inspection of the
ScatterNet's patterns reveal that they may be less useful for discriminative
tasks, and we believe this may be causing the current gaps in state-of-the-art
performance between the two. 

We have presented an architectural change to ScatterNets that can make it
sensitive to more recognizable shapes. We believe that using this new layer is
how we can start to close the gap, making more generic and descriptive
ScatterNets while keeping control of their desirable properties. 

A future paper will include classifier results for these new filters.

\bibliographystyle{IEEEbib}
\bibliography{MyLibrary}

\end{document}